\documentclass[conference]{IEEEtran}
\IEEEoverridecommandlockouts
\usepackage{cite}
\usepackage{amsmath,amssymb,amsfonts}
\usepackage{algorithmic}
\usepackage{graphicx}
\usepackage{textcomp}
\usepackage{xcolor}
\usepackage{bm}
\usepackage{balance}
\usepackage{siunitx}
\usepackage{subcaption} 
\newtheorem{mydef}{Definition}
\newtheorem{theorem}{Theorem}

\def\BibTeX{{\rm B\kern-.05em{\sc i\kern-.025em b}\kern-.08em
    T\kern-.1667em\lower.7ex\hbox{E}\kern-.125emX}}
\begin{document}

\title{Whole-body Dynamic Collision Avoidance with Time-varying Control Barrier Functions}

\author{Jihao Huang, Xuemin Chi, Zhitao Liu$^{\dagger}$, Hongye Su%
\thanks{The authors would like to express appreciation for the financial support provided  by National Key R\&D Program of China (Grant NO. 2021YFB3301000); National Natural Science Foundation of China (NSFC:62173297), Zhejiang Key R\&D Program (Grant NO. 2022C01035); Project of Ningbo Automotive Electronics Intelligentification Innovation Union (2022H007).
}
\thanks{
Jihao Huang, Xuemin Chi, Zhitao Liu and Hongye Su are with the State Key Laboratory of Industrial Control Technology, Institute of Cyber-Systems and Control, Zhejiang University, Hangzhou, China {\tt\footnotesize \{jihaoh, chixuemin, ztliu, hysu\}@zju.edu.cn}.
}%
\thanks{$^\dagger$ Corresponding author.}
}

\maketitle
\begin{abstract}
Recently, there has been increasing attention in robot research towards the whole-body collision avoidance.
In this paper, we propose a safety-critical controller that utilizes time-varying control barrier functions (time varying CBFs) constructed by Robo-centric Euclidean Signed Distance Field (RC-ESDF) to achieve dynamic collision avoidance.
The RC-ESDF is constructed in the robot body frame and solely relies on the robot's shape, eliminating the need for real-time updates to save computational resources.
Additionally, we design two control Lyapunov functions (CLFs) to ensure that the robot can reach its destination.
To enable real-time application, our safety-critical controller which incorporates CLFs and CBFs as constraints is formulated as a quadratic program (QP) optimization problem.
We conducted numerical simulations on two different dynamics of an L-shaped robot to verify the effectiveness of our proposed approach.
\end{abstract}

\section{Introduction}
\label{sec:introduction}
Collision avoidance is a crucial technology in robotics~\cite{alonso2018cooperative}. 
Recently, as robots are being increasingly utilized in various fields such as autonomous driving, military operations, delivery services, and medical procedures, whole-body collision avoidance has been gaining increasing attention.
Whole-body collision avoidance considers the shape of the robot instead of treating it as a point mass.
However, achieving real-time whole-body collision avoidance with any-shaped robot and obstacles remains a significant challenge.
In addition, control barrier function (CBF) has recently been widely employed for safety-critical control which prioritize safety aspects such as collision avoidance and physical limits over navigation and tracking~\cite{huang2023obstacle}.

Researchers have developed various approaches to tackle the issue of whole-body collision, including Gilbert-JohnsonKeerthi (GJK)-based methods~\cite{gilbert1988fast,zhang2020optimization}, corridor-based techniques~\cite{han2021fast,li2021optimal,wang2022geometrically}, Euclidean Signed Distance Field (ESDF)-based approaches~\cite{oleynikova2016signed, geng2023robo}, and specific approaches~\cite{wang2023linear}.
GJK-based methods require complex pre-processing for computing the distance between two convex objects.
Corridor-based techniques create a sequence of convex regions in the free space to ensure the robot safety by confining the robot within safe corridors.
However, corridor-based techniques necessitate that at least one robot be present in the intersection of any two adjacent safe corridors; if this condition is not met, no feasible solutions can be obtained.
ESDF-based approaches involve additional computational and memory overhead, few approaches can satisfy both accuracy and low computational consumption.
Robo-centric ESDF (RC-ESDF) is proposed in~\cite{geng2023robo} to solve these issues with low computational overhead and accurate collision detection, however, this method has not been extended to scenarios involving dynamic obstacles.
Scale optimization is employed in \cite{wang2023linear} to enable collision avoidance between two polytopes. Nevertheless, this method is unsuitable for non-convex shapes of robots and obstacles.

A safety-critical controller~\cite{ames2014control, ames2016control} which incorporates control Lyapunov function (CLF) for stability and CBF for safety are proposed to achieve adaptive cruise control (ACC).
Since both stability and safety constraints are affine in the optimize variables and the entire optimization problem is organized by a quadratic program (QP), real-time optimal solutions can be obtained.
Furthermore, this paradigm has also been successfully extended to robotics for navigation and collision avoidance~\cite{wu2016safety, zeng2021safety, thirugnanam2022safety, huang2023obstacle}, as well as various approaches~\cite{xiao2019control, xiao2023barriernet}.
Dynamic collision avoidance with time-varying control barrier function (time-varying CBF) is proposed in~\cite{jian2023dynamic,huang2023obstacle}.
However, both the robot and obstacles in these works have circular shapes which are not easily applicable to robots and obstacles of other shapes.
Collision avoidance between polytopes based on CBF and duality theorem is proposed in~\cite{thirugnanam2022safety}, however, this approach does not consider dynamic collision avoidance and may not provide real-time solutions when there are a large number of obstacles.

Taking into account all the issues mentioned above, our work proposes constructing time-varying CBFs with RC-ESDF to achieve whole-body dynamic collision avoidance between two arbitrarily shaped robots and obstacles, and optimal solutions can be obtained in real-time.
In our approach, obstacles are sampled as multiple collision points, and each collision point constructs a CBF with RC-ESDF in the robot body frame.
Additionally, our work designs two CLFs to navigate the robot to its destination.
Since all safety and stability constraints are affine in the control variable, real-time solutions are obtainable regardless of the number of constraints.
We validate the effectiveness of our method through numerical simulations involving an L-shaped robot and multiple static and dynamic obstacles.

The paper is structured as follows:
Sec.~\ref{sec:background} presents the problem formulation and background information on CLF, CBF, and RC-ESDF.
Sec.~\ref{sec:method} presents the detailed design of CLFs and CBF with the controller synthesis.
In Sec.~\ref{sec:experiment}, numerical simulations are used to verify the effectiveness of our proposed approach.
Concluding remarks are provided in Sec.~\ref{sec:conclusion}.

\section{Background}
\label{sec:background}
In this section, we provide relevant background information to provide more details about our work.
Firstly, we define the problem formulation and introduce the dynamics of the robot used in our study.
Additionally, we introduce some important concepts: control Lyapunov function (CLF), control barrier function (CBF), and Robo-centric ESDF (RC-ESDF).

\subsection{Problem Formulation}
In this work, our main goal is to navigate a robot to its destination while ensuring it avoids colliding with both static and dynamic obstacles.
We consider two dynamics of the robot in this work. 
The first one is the single integral model as follows:
\begin{equation}
    \left[\begin{array}{c}
    \dot{x} \quad \dot{y} 
    \end{array}\right]^T = 
    \left[\begin{array}{c}
    v_x \quad v_y
    \end{array}\right]^T,
    \label{eq:single_integral_model}
\end{equation}
where $\bm{p}= [x, y]^T$ represents the position of the robot and $[v_x, v_y]^T$ represents the velocity which controls the movements of robot.
The second model is the unicycle model as follows:
\begin{equation} 
    \left[\begin{array}{c}
    \dot{x} \quad
    \dot{y} \quad
    \dot{\theta}
    \end{array}\right]^T = 
    \left[\begin{array}{c}
    v \cos \theta \quad
    v \sin \theta \quad
    \omega
    \end{array}\right]^T,
    \label{eq:unicycle_model}
\end{equation}
where $\theta$ represents the orientation of the robot with respect to the $x$ axis, $v$ and $\omega$ represent the linear and angular velocity, respectively, which control the robot's motion.
The dynamics of all obstacles follow a double integral model
\begin{equation} 
    \left[\begin{array}{c}
    \dot{x}_o \quad
    \dot{y}_o \quad
    \dot{v}_{ox} \quad
    \dot{v}_{oy}
    \end{array}\right]^T = 
    \left[\begin{array}{c}
    v_{ox} \quad
    v_{oy} \quad
    a_{ox} \quad
    a_{ox}
    \end{array}\right]^T,
    \label{eq:dynamics_obs}
\end{equation}

\subsection{Control Lyapunov Function}
Control Lyapunov function (CLF) is proposed to achieve the control objectives of stabilizing a system to an equilibrium point.
Suppose there is a control affine system
\begin{equation}
    \dot{\bm{x}} = f(\bm{x}) + g(\bm{x})\bm{u},
    \label{eq:system}
\end{equation}
for $\bm{x} \in \mathcal{D}$, where $\mathcal{D} \subset \mathbb{R}^n$ and $\bm{u} \in \mathcal{U}$, where $\mathcal{U}$ denotes the set of admissible inputs, defined as
\begin{equation}
    \mathcal{U}:=\{\bm{u} \in \mathbb{R}^m, \bm{u}_{\text{min}} \leq \bm{u} \leq \bm{u}_{\text{max}}\},
\end{equation}
where $\bm{u}_{\text{min}}$ and $\bm{u}_{\text{max}}$ represent the lower and upper bounds of $\bm{u}$.
Moreover, the mapping $f:\mathcal{D} \to \mathbb{R}^n$ and $g:\mathcal{D} \to \mathbb{R}^{n \times m}$ are locally Lipschitz continuous on $\mathcal{D}$.
The single integral model \eqref{eq:single_integral_model} and the unicycle model \eqref{eq:unicycle_model} can be converted to the form of \eqref{eq:system}, taking \eqref{eq:unicycle_model} as an example:
\begin{equation} 
    \left[\begin{array}{c}
    \dot{x} \\
    \dot{y} \\
    \dot{\theta} 
    \end{array}\right] = 
    \left[\begin{array}{c}
    0 \\
    0 \\
    0
    \end{array}\right] + 
    \left[\begin{array}{cc}
    \cos\theta & 0 \\
    \sin\theta & 0 \\
    0 & 1
    \end{array}\right]
    \left[\begin{array}{c}
    v \\
    \omega
    \end{array}\right]
\end{equation}
where $\bm{x}=[x, y, \theta]^T \in \mathbb{R}^3$ and $\bm{u}=[v, \omega]^T \in \mathbb{R}^2$.
Before introducing the concept of CLF, we first introduce the concept of class $\mathcal{K}$ function $\alpha(\cdot)$: $[0, a) \to [0, \infty)$ is said to belong to class $\mathcal{K}$ function if it is strictly increasing and satisfies $\alpha(0) = 0$.
Furthermore, it is said to belong to class $\mathcal{K}_{\infty}$ function if it belong to class $\mathcal{K}$ function and also satisfies $a = \infty$ and $\alpha(b) \to \infty$ as $b \to \infty$.
\begin{mydef}
    A continuously differential function $V: \mathcal{D} \to \mathbb{R}$ is a control Lyapunov function (CLF) for system \eqref{eq:system} if it is positive definite and satisfies~\cite{ames2019control}:
    \begin{equation}
        \inf_{\bm{u} \in \mathcal{U}} [L_f V(\bm{x}) + L_g V(\bm{x}) \bm{u}] \leq -\gamma(V(\bm{x})),
        \label{eq:define_clf}
    \end{equation}
    where $L_f V(\bm{x})=\frac{\partial V}{\partial \bm{x}} f(\bm{x})$ and $L_g V(\bm{x})=\frac{\partial V}{\partial \bm{x}}g(\bm{x})$ are Lie-derivatives of $V(\mathbf{x})$, $\gamma(\cdot)$ belongs to the class $\mathcal{K}$ function.
    \label{def:1}
\end{mydef}
\begin{theorem}
Given a CLF $V(\bm{x})$ defined in Def.~\ref{def:1}, any Lipschitz continuous controller $\bm{u} \in K_{\text{clf}}(\bm{x})$, with 
\begin{equation}
    K_{\text{clf}} (\bm{x}):=\{\bm{u} \in \mathcal{U}, L_f V(\bm{x}) + L_g V(\bm{x}) \bm{u} \leq -\gamma(V(\bm{x})) \},
    \label{eq:kclf}
\end{equation}
can stabilize the system \eqref{eq:system} to the origin.
\end{theorem}

Therefore, we can use the affine constraint \eqref{eq:kclf} in $\bm{u}$ to formulate an optimization-based controller.

\subsection{Control Barrier Function}
Compared with CLF which leads the system to an equilibrium point, control barrier function (CBF) are proposed to ensure safety by enforcing the invariance of a set, i.e., not leaving the safe set.
Consider a set $\mathcal{C} \subset \mathcal{D}$ defined as a zero-superlevel set of a continuously differentiable function $h$: $\mathcal{D} \to \mathbb{R}$, yielding:
\begin{equation}
\begin{aligned}
    \mathcal{C} & = \{\bm{x} \in \mathcal{D} \subset \mathbb{R}^n : h(\bm{x}) \geq 0 \}, \\
    \partial \mathcal{C} & = \{\bm{x} \in \mathcal{D} \subset \mathbb{R}^n : h(\bm{x}) = 0 \}, \\
    \rm Int(\mathcal{C}) & = \{\bm{x} \in \mathcal{D} \subset \mathbb{R}^n : h(\bm{x}) > 0 \}.
    \label{eq:safe_set}
\end{aligned} 
\end{equation}
We refer to $\mathcal{C}$ as the safe set.
\begin{mydef}
    A set $\mathcal{C} \subset \mathcal{D}$ is forward invariant w.r.t \eqref{eq:system} if for each $\bm{x}_0 \in \mathcal{C}, \bm{x}(t) \in \mathcal{C}$ for $\bm{x}(0) = \bm{x}_0, \forall t \geq 0$.
    The system \eqref{eq:system} is safe with respect to the safe set $\mathcal{C}$ if $\mathcal{C}$ is forward invariant.\label{def:2}
\end{mydef}
\begin{mydef}
    Given a set $\mathcal{C}$ defined by \eqref{eq:safe_set}, the continuously differentiable function $h$: $\mathcal{D} \to \mathbb{R}$ is called the control barrier function (CBF) if $\frac{\partial h(\bm{x})}{\partial \bm{x}} \not=0, \forall \bm{x} \in \partial \mathcal{C}$ and there exists an extended class $\mathcal{K}_{\infty}$ function $\alpha(\cdot)$ such that~\cite{ames2019control}
    \begin{equation}
        \sup_{\bm{u} \in \mathcal{U}} [L_f h(\bm{x}) + L_g h(\bm{x})\bm{u}] \geq -\alpha(h(\bm{x})),
        \label{eq:define_cbf}
    \end{equation}
    where $L_f h(\bm{x}) = \frac{\partial h(\bm{x})}{\partial \bm{x}}f(\bm{x})$ and $L_g h(\bm{x}) = \frac{\partial h(\bm{x})}{\partial \bm{x}}g(\bm{x})$ are Lie-derivatives of $h(\bm{x})$.
    \label{def:3}
\end{mydef}
\begin{theorem}
    If $h$ is a CBF associated with a set $\mathcal{C}$ and $\frac{\partial h(\bm{x})}{\partial \bm{x}} \not= 0, \forall \bm{x} \in \partial \mathcal{C}$, then any Lipschitz continuous controller $\bm{u} \in K_{\text{cbf}}$, with 
    \begin{equation}
    K_{\text{cbf}}(\bm{x}):=\{\bm{u} \in \mathcal{U}, L_f h(\bm{x}) + L_g h(\bm{x})\bm{u} \geq -\alpha(h(\bm{x}))\}.
    \label{eq:kcbf}
    \end{equation}
    can guarantee the forward invariance of the set $\mathcal{C}$ and thus the safety of system \eqref{eq:system}.
\end{theorem}

The constraint \eqref{eq:kcbf} is also an affine constraint in $\bm{u}$, so it can be combined with \eqref{eq:kclf} to formulate a quadratic optimization problem.
Furthermore, to simplify the whole optimization problem, we choose constant scalars as the class $\mathcal{K}$ function for both CLFs and CBFs, i.e., $\gamma(V(\bm{x})) = \gamma V(\bm{x})$, $\alpha(h(\bm{x})) = \alpha h(\bm{x})$, where $\gamma$ and $\alpha$ are constant scalars.

\subsection{Robo-Centric Euclidean Signed Distance Field}
\label{sec:rc-esdf}
\begin{figure}
    \centering
    \includegraphics[width=0.80\linewidth]{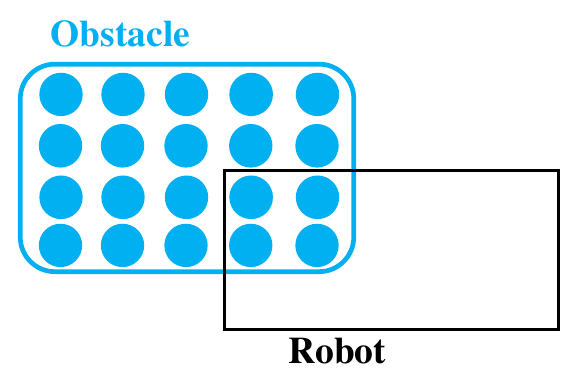}
    \caption{The entire obstacle is represented by multiple collision points, which are shown as blue dots. The robot needs to avoid colliding with all of these points.}
    \label{fig:collision_point}
\end{figure}
In this section, we present the construction details of the Robo-centric Euclidean Signed Distance Field (RC-ESDF).
The RC-ESDF is defined in the robot's body frame and relies solely on the robot's shape, disregarding obstacles' shapes~\cite{geng2023robo}.
Compared to the original ESDF based approaches, the RC-ESDF based approach is not required to update based on the real-time environment information from the onboard sensors, which can greatly reduce computation costs.
Assume there are $N$ static and dynamic obstacles in the environment and each obstacle is represented by $\text{O}_i \in \mathbb{O}=\{\text{O}_0, \text{O}_1, \dots, \text{O}_{N-1}\}$.
Each obstacle $\text{O}_i$ is sampled as $M_i$ collision points, as shown in Fig.~\ref{fig:collision_point}.
The robot must ensure collision avoidance with all the collision points of each obstacle, meaning that all collision points should be outside or on the surface of the robot.
Hence, we define that when a collision point lies within the robot, its RC-ESDF value is negative and its norm represents the closest distance to the robot's surface.
Contrarily, its RC-ESDF value is greater than zero.
Therefore, if all collision points have RC-ESDF values greater than zero, collision avoidance can be achieved.
Hence, we utilized RC-ESDF to build CBFs for achieving collision avoidance in our work, more details can refer to Sec.~\ref{sec:design_cbf}.

\section{Controller Design}
\label{sec:method}
This section will show how to design a controller which guides the robot to its destination while avoiding all obstacles.
Firstly, we will demonstrate the design of CLFs for achieving navigation.
Additionally, we will explain how to design CBFs using RC-ESDF.
Lastly, we will illustrate how to combine the CLFs and CBFs to formulate a safety-critical controller.

\subsection{Design of Control Lyapunov Function}
In this section, we will demonstrate how to design CLFs for guiding the robot to its destination, and we have designed two CLFs for this purpose.
Assume that the destination of the robot is represented by $(x_\text{d}, y_\text{d}, \theta_\text{d})$.
The first CLF is designed to reduce the distance between the current position and the destination,
\begin{equation}
    V_\text{d}(\bm{x}) = (x - x_\text{d})^2 + (y - y_\text{d})^2,
    \label{eq:v1}
\end{equation}
and we can get $\frac{\partial V_\text{d}(\bm{x})}{\partial \bm{p}} = [2(x - x_\text{d}), 2(y - y_\text{d})]^T$.
With the robot dynamics \eqref{eq:single_integral_model} or \eqref{eq:unicycle_model}, we can obtain $L_f V_\text{d}(\bm{x}) = 0$.
Furthermore, a robot with the single integral model \eqref{eq:single_integral_model} can solely rely on this CLF $V_\text{d}(\bm{x})$ to reach its destination.
However, for the robot with the unicycle model \eqref{eq:unicycle_model}, we have $L_g V_\text{d}(\bm{x}) = [2(x - x_\text{d})\cos\theta + 2(y - y_\text{d})\sin\theta, 0]$, it can only control the linear velocity $v$ for navigation.
Hence, when the robot needs to rotate to reach its destination, it is not sufficient to rely solely on $V_d(\bm{x})$ for navigation.
Therefore, we design another CLF $V_\theta(\bm{x})$, which adjusts $\omega$ to guide the robot towards its destination.
\begin{equation}
    V_\theta(\bm{x}) = [\cos\theta(y_\text{d} - y) - \sin\theta(x_\text{d} - x)]^2,
    \label{eq:v2}
\end{equation}
and we have
\begin{equation}
    \frac{\partial V_\theta(\bm{x})}{\partial \bm{x}} = 2t_2
    \left[\begin{array}{c}
    \sin \theta \\
    -\cos \theta \\
    -\sin\theta(y_\text{d} - y) - \cos\theta(x_\text{d} - x)
    \end{array}\right]
\end{equation}
where $t_2 = 2[\cos\theta(y_\text{d} - y) - \sin\theta(x_\text{d} - x)]$.
For the robot with the unicycle model \eqref{eq:unicycle_model}, $L_f V_\theta(\bm{x}) = 0$ and $L_g V_\theta(\bm{x}) = [0, -\sin\theta(y_\text{d} - y) - \cos\theta(x_\text{d} - x)]$, hence $V_\theta(\bm{x})$ can only adjust $\omega$ for navigation.
In conclusion, for the robot system \eqref{eq:single_integral_model}, only $V_\text{d}(\bm{x})$ can guide the robot to its destination. 
On the other hand, for the system \eqref{eq:unicycle_model}, both $V_\text{d}(\bm{x})$ and $V_\theta(\bm{x})$ need to be combined in order to control both $v$ and $\omega$ and achieve navigation.

\subsection{Design of Control Barrier Functions}
\label{sec:design_cbf}
In this section, we will demonstrate how to design CBFs and formulate the constraints of CBFs based on RC-ESDF in order to avoid collisions with both static and dynamic obstacles.
When the robot is at state $\bm{x} = [\bm{p}, \theta]^T$, we consider using $h_{i, j}$ to represent the RC-ESDF value of the $j$th collision point of obstacle $\text{O}_i$ at point $\bm{q}_{b,i}^{j}$ in the robot body frame, and we have a corresponding transformation relationship:
\begin{equation}
    \bm{q}_{b,i}^{j} = \bm{R}(\theta)^{-1}(\bm{q}_{w,i}^{j} - \bm{p}),
    \label{eq:transform}
\end{equation}
where $\bm{q}_{w,i}^{j}$ represents the coordinate of this collision point in the world frame and 
\begin{equation}
    \bm{R}(\theta) = \left[\begin{array}{cc}
    \cos\theta & -\sin\theta \\
    \sin\theta & \cos\theta
    \end{array}\right].
\end{equation}
As we mentioned in Sec.~\ref{sec:rc-esdf}, if $h_{i, j} > 0$, then we can ensure that the robot avoids colliding with the $j$th collision point of $\text{O}_i$.
Therefore, we consider using the RC-ESDF value $h_{i, j}$ as a CBF, which also needs to satisfy the constraint \eqref{eq:define_cbf}.
It is important to mention that in the presence of dynamic obstacles in the environment, time-varying control barrier functions (time-varying CBFs) should be employed to prevent collisions with these obstacles, which impose stricter constraint on the controller~\cite{huang2023obstacle}
\begin{equation}
    \sup_{\bm{u} \in \mathcal{U}} [L_f h_{i, j} (\bm{x}, t) + L_g h_{i,j} (\bm{x}, t) \bm{u} + \frac{\partial h_{i, j}(\bm{x}, t)}{\partial t}] \geq -\alpha(h_{i,j}(\bm{x}, t)),
    \label{eq:define_tvcbf}
\end{equation}
where $\frac{\partial h_{i,j}(\bm{x}, t)}{\partial t} = \frac{\partial h_{i,j}(\bm{x}, t)}{\partial \bm{q}_{w,i}^{j}(t)} \frac{\partial \bm{q}_{w,i}^{j}(t)}{\partial t}$ shows the influence of the dynamic obstacle's position.
For static obstacles, $\frac{\partial \bm{q}_{w,i}^{j}(t)}{\partial t} = 0$.
In addition, the set of controller \eqref{eq:kcbf} which renders the $\mathcal{C}$ corresponding to $h_{i,j}(\bm{x}, t)$ safe is converted to
\begin{equation}
\begin{aligned}
    K_{\text{cbf}}(\bm{x}):=\{\bm{u} \in \mathcal{U}, L_f h_{i, j}(\bm{x}, t) + L_g & h_{i, j}(\bm{x}, t) \bm{u} + \frac{\partial h_{i, j}(\bm{x}, t)}{\partial t} \\
    & \geq -\alpha(h_{i, j}(\bm{x}, t)) \}.
    \label{eq:ktvcbf}
\end{aligned}
\end{equation}

Since we have $L_g h_{i, j} = \frac{\partial h_{i, j}}{\partial \bm{x}}g(\bm{x})$, where $\frac{\partial h_{i, j}}{\partial \bm{x}} = [\frac{\partial h_{i, j}}{\partial \bm{p}}, \frac{\partial h_{i, j}}{\partial \theta}]^T$ and the transformation relationship \eqref{eq:transform}.
Hence the gradients of $h_{i, j}$ with respect to $\bm{p}$ and $\theta$ can be calculated as follows:
\begin{equation}
\begin{aligned}
    \frac{\partial h_{i, j}}{\partial \bm{p}} &= \frac{\partial \bm{q}_{b,i}^{j}}{\partial \bm{p}} \frac{\partial h_{i, j}}{\partial \mathbf{q}_{b,i}^{j}}
    = -\mathbf{R}(\theta) \frac{\partial h_{i, j}}{\partial \mathbf{q}_{b,i}^{j}}, \\
    \frac{\partial h_{i, j}}{\partial \theta} &= (\frac{\partial \bm{q}_{b,i}^{j}}{\partial \theta})^T \frac{\partial h_{i, j}}{\partial \bm{q}_{b,i}^{j}} \\
    &= (\bm{q}_{w,i}^{j} - \bm{p})^T
    \left[\begin{array}{cc}
    -\sin\theta & -\cos\theta \\
     \cos\theta & -\sin\theta
    \end{array}\right] \frac{\partial h_{i, j}}{\partial \bm{q}_{b,i}^j}.
\end{aligned}
\end{equation}
In addition, the gradient of $h_{i, j}$ with respect to $\bm{q}_{w,i}^j$ 
\begin{equation}
    \frac{dh_{i, j}}{d\bm{q}_{w,i}^j} = \frac{d\bm{q}_{b,i}^j}{d\bm{q}_{w,i}^j} \frac{dh_{i, j}}{d\bm{q}_{b,i}^j}
    = \bm{R}(\theta) \frac{dh_{i, j}}{d\bm{q}_{b,i}^j}.
\end{equation}

As we mentioned above, the RC-ESDF value is solely determined by the robot's shape, and different shapes of robots require different calculation methods.
Taking a circular-shaped robot with radius $r$ as an example, we can obtain
\begin{equation}
    h_{i, j} = \| \bm{q}_{b,i}^j \|_2 - r, \quad
    \frac{\partial h_{i, j}}{\partial \bm{q}_{b,i}^j} = \frac{\bm{q}_{b,i}^j}{\| \bm{q}_{b,i}^j \|_2},
\end{equation}
where the gradient of $h_{i, j}$ with respect to $\bm{q}_{b,i}^j$ can be calculated analytically.
However, for the rectangle-shaped robot with half length $l_r$ and half weight $w_r$, we can obtain
\begin{equation}
\begin{aligned}
    dx &= \mathbf{q}_{b,i}^{j, x} - l_r, \quad dy = \mathbf{q}_{b,i}^{j, y} - w_r, \\
    d_{\text{in}} &= \text{min}(\text{max}(dx, dy), 0), \\
    d_{\text{out}} &= (\text{max}(dx, 0))^2 + (\text{max}(dy, 0))^2, \\
    h_{i, j} & = d_{\text{in}} + d_{\text{out}}.
\end{aligned}
\end{equation}
Moreover, in this case the gradient of $h_{i, j}$ with respect to $\bm{q}_{b,i}^j$ can not be calculated analytically due to the discontinuity.
Thus the gradient can be calculated numerically as follows:
\begin{equation}
    \frac{\partial h_{i, j}}{\partial \bm{q}_{b,i}^j} = \frac{h_{i, j}(\bm{q}_{b,i}^j + \Delta \bm{q}) - h_{i, j}(\bm{q}_{b,i}^j)}{\Delta \bm{q}}
\end{equation}
where $\Delta \bm{q}$ is a small amount.

In conclusion, robot needs to avoid collisions with all collision points $\bm{q}_{w,i}^j, i \in \{0, 1, \dots, N-1\}, j \in \{0, 1, \dots, M_i - 1\}$.
Thus we can formulate the CBFs $h_{i, j}$ with respect all these collision points and add these constraints to the optimization problem to ensure collision avoidance with all obstacles.

\subsection{Design of Controller}
Since both the constraints \eqref{eq:kclf} for CLF and \eqref{eq:ktvcbf} for CBF have affine forms in $\bm{u}$, we can combine them and formulate a quadratic optimization problem to get real-time solutions.
In this work, we propose a safety-critical controller formulation based on Quadratic Program (QP) that combines CBFs for safety and CLFs for stability as follows:
\noindent\rule{\columnwidth}{0.8pt}
\textbf{CLF-CBF-QP:}
\begin{subequations}
\begin{align}
    \min_{(\bm{u}, \bm{\delta}) \in \mathbb{R}^{m + 2}} & \frac{1}{2}\bm{u}^T R \bm{u} + \bm{\delta}^T H \bm{\delta} \label{eq:problem} \\
    \text{s.t.} ~& L_f V_\text{d}(\bm{x}) + L_g V_\text{d}(\bm{x}) \bm{u} + \gamma_\text{d}(V_\text{d}(\bm{x})) \leq \delta_\text{d}, \label{eq:cons_distance_clf} \\ 
    & L_f V_\theta(\bm{x}) + L_g V_\theta(\bm{x}) \bm{u} + \gamma_\theta(V_\theta(\bm{x})) \leq \delta_\theta, \label{eq:cons_theta_clf} \\ \notag
    & L_f h_{i, j}(\bm{x}, t) + L_g h_{i, j}(\bm{x}, t) \bm{u} + \frac{\partial h_{i, j}(\bm{x}, t)}{\partial t} \\ \notag
    & \quad + \alpha(h_{i, j}(\bm{x}, t)) \geq 0, i=0, 1, \dots, N-1,  \\
    & \quad \quad \quad \quad \quad \quad \quad \quad j=0, 1, \dots, M_i - 1, \label{eq:cons_cbf} \\
    & \bm{u} \in \mathcal{U}, \label{eq:cons_u} \\
    & \bm{\delta} \in \mathbb{R}^2. \label{eq:cons_delta} 
\end{align}
\label{eq:optimal_problem}
\end{subequations}
\noindent\rule{\columnwidth}{0.4pt}
\noindent
where $\bm{u} \in \mathbb{R}^m$ is the control variable and $\bm{\delta} = [\delta_\text{d}, \delta_\theta]^T \in \mathbb{R}^2$ is the slack variable for CLFs.
Furthermore, $R$ and $H$ are positive definite matrices, where $R$ is the weight matrix for $\bm{u}$ and $H$ is the weight matrix for $\bm{\delta}$.
The objective function \eqref{eq:problem} of this quadratic optimization problem can be divided into two parts:
The first part aims to minimize $\bm{u}$ as much as possible, while the second part limits the additional cost of slack variable $\bm{\delta}$.
The constraints \eqref{eq:cons_distance_clf} and \eqref{eq:cons_theta_clf} of CLF are relaxed by introducing the slack variable $\bm{\delta}$, which means that when CLF constraints conflict with CBF constraints (\eqref{eq:cons_cbf}), the safety-critical controller must relax constraints of stability to guarantee safety.
Safety constraints are imposed by \eqref{eq:cons_cbf}, ensuring collision avoidance with all collision points of each obstacle.
Constraint \eqref{eq:cons_u} restricts control variable $\bm{u}$ within a permissible range to satisfy robot's dynamics constraint.
Additionally, \eqref{eq:optimal_problem} is designed for the robot with unicycle model \eqref{eq:unicycle_model}.
When the robot with the single integral model \eqref{eq:single_integral_model}, constraint \eqref{eq:cons_theta_clf} can be omitted.

\section{Numerical validation}
\label{sec:experiment}
\begin{table}
    \caption{Setup of the simulation parameters}
    \label{tab:simulation_params}
    \centering
    \begin{tabular}{l|l|l}
    \hline
    Notation & Meaning & Value     \\ \hline
    $\alpha$ & Class $\mathcal{K}$ function for control barrier functions & 1.0 \\
    $\gamma_\text{d}$ & Class $\mathcal{K}$ function for CLF $V_\text{d}(\bm{x})$ & 1.0 \\
    $\gamma_\theta$   & Class $\mathcal{K}$ function for CLF $V_\theta(\bm{x})$   & 3.0 \\
    $p$               & Diagonal matrix elements for $H$                          & 1000 \\
    $v_x^{\text{max}}$ & Robot's maximum horizontal velocity & $2.0 \, \si[per-mode=symbol]{\metre\per\second}$ \\
    $v_y^{\text{max}}$ & Robot's maximum vertical velocity & $2.0 \, \si[per-mode=symbol]{\metre\per\second}$ \\
    $v_{\text{max}}$ & Robot's maximum linear velocity & $2.0 \, \si[per-mode=symbol]{\metre\per\second}$ \\
    $w_{\text{max}}$ & Robot's maximum angular velocity & $1.0 \, \si[per-mode=symbol]{\radian\per\second}$ \\
    $M$              & Number of sampled collision points & 24 \\
    \hline     \end{tabular}%
\end{table}

The effectiveness of our method is validated through numerical simulations.
The simulations are conducted on an Ubuntu Laptop with Intel Core i9-13900HX processor using Python for all computations.
The quadratic optimization problem \eqref{eq:optimal_problem} is solved by QPOASES for real-time application.
L-shaped robot is employed in the numerical simulations, and the RC-ESDF value is calculated by splitting the L-shaped robot into two rectangular robots.
The time step $\Delta t$ of simulation is set as $0.1 \, \si[per-mode=symbol]{\second}$ and the maximum loop time $t_{\text{max}}$ is set as $20\, \si[per-mode=symbol]{\second}$.
Moreover, $v_x^\text{max} = -v_x^\text{min}$ and $v_y^\text{max} = -v_y^\text{min}$ represent the maximum of the horizontal and vertical velocity respectively for the single integral model \eqref{eq:single_integral_model}.
And $v_\text{max} = -v_\text{min}$ and $w_\text{max} = -w_\text{min}$ represent the maximum of linear velocity and angular velocity for the unicycle model \eqref{eq:unicycle_model}.
The number $M$ of sampled collision points of the polytopic-shaped robot is chosen as 24.
We have presented the parameters for both the robot dynamics and the CLF-CBF-QP controller in Tab.~\ref{tab:simulation_params}.
In the following, we will present simulation results for two different robot dynamics across various scenarios.

\subsection{Robot with Single Integral Model}
\begin{figure}
    \centering
    \begin{subfigure}{0.49\linewidth}
        \centering
        \includegraphics[width=0.98\linewidth]{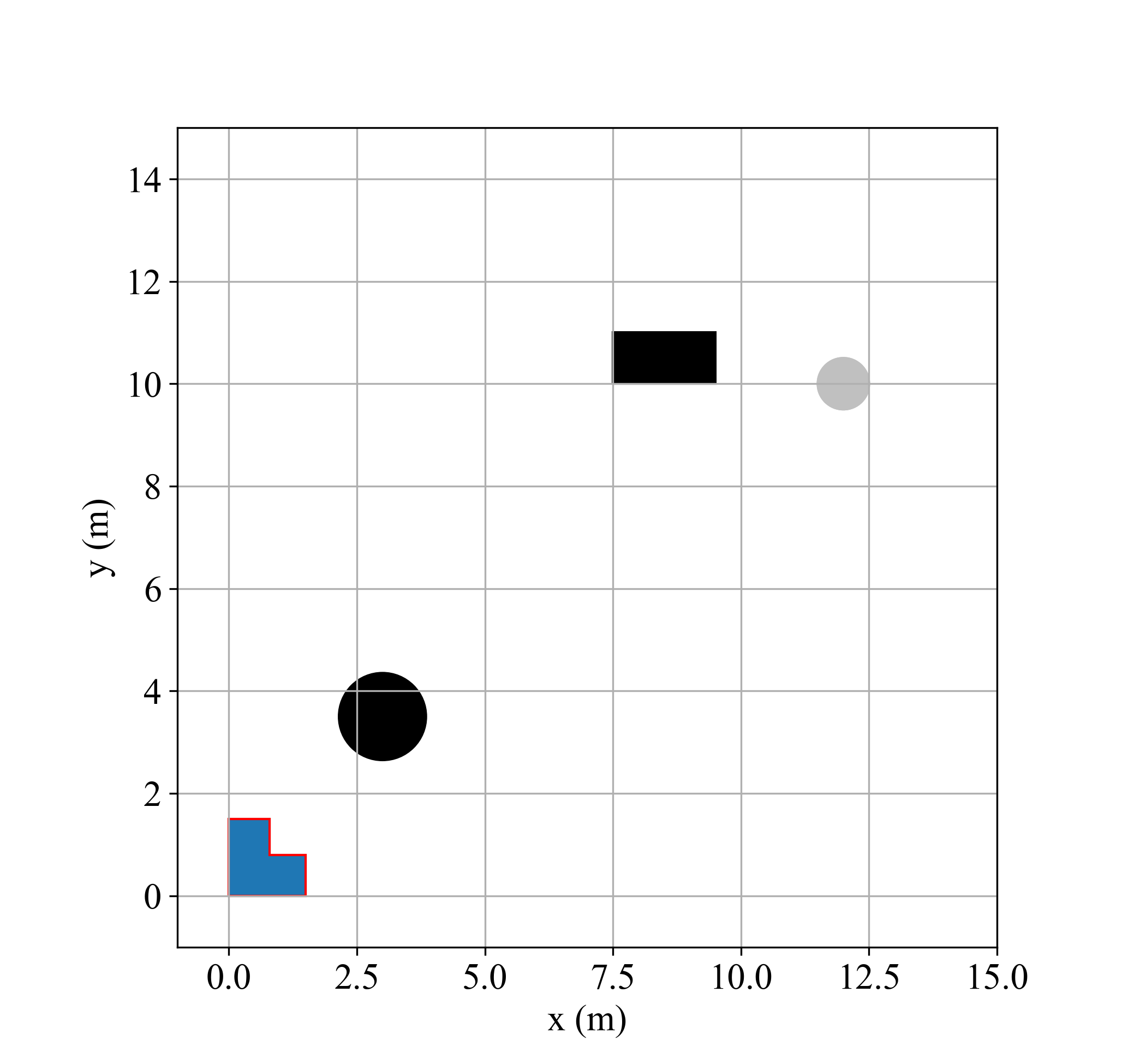}
        \caption{$t = 0.0 \, \si[per-mode=symbol]{\second}$}
        \label{subfig:integral_case1}
    \end{subfigure}
    \centering
    \begin{subfigure}{0.49\linewidth}
        \centering
        \includegraphics[width=0.98\linewidth]{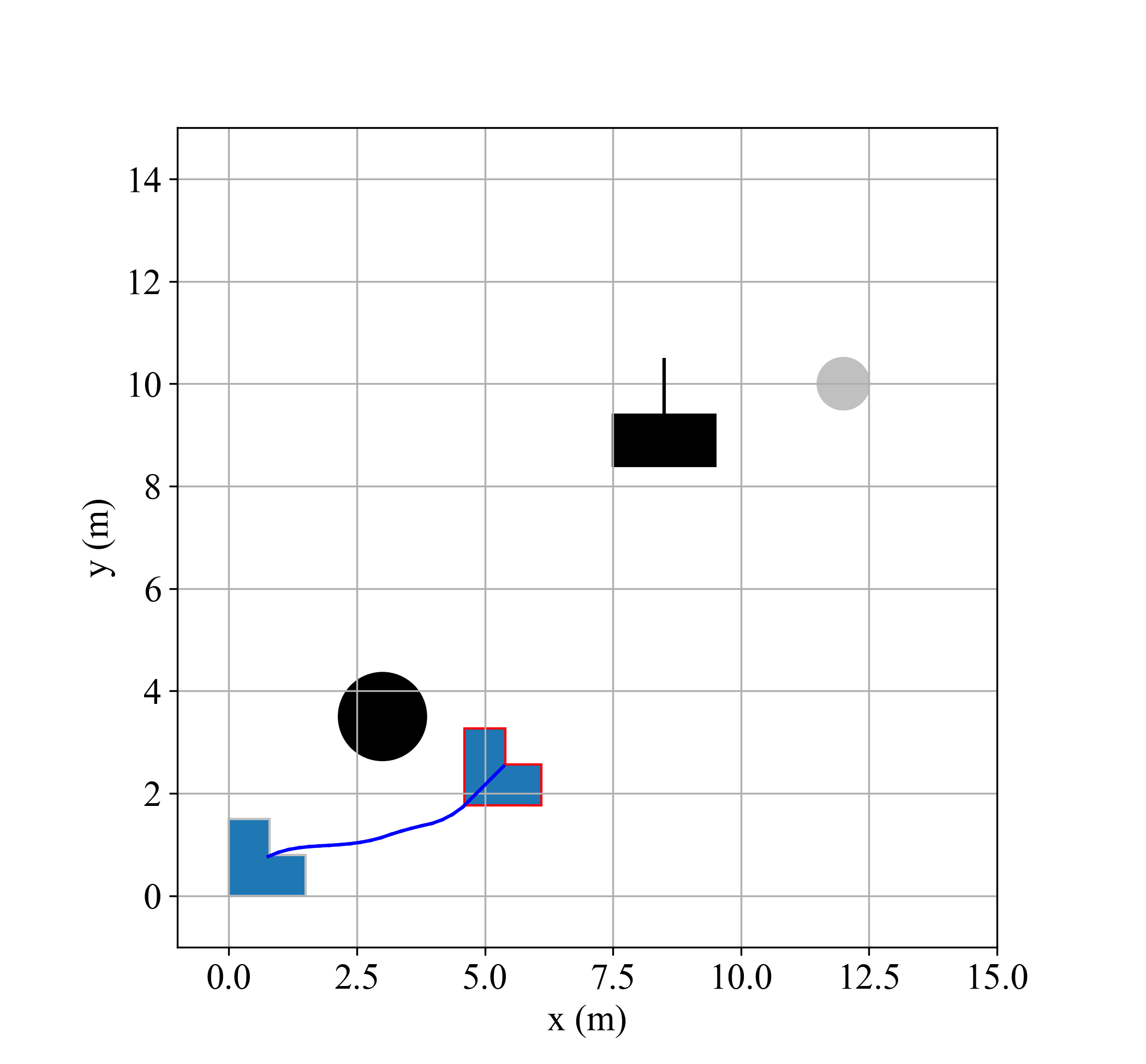}
        \caption{$t = 2.3 \, \si[per-mode=symbol]{\second}$}
        \label{subfig:integral_case2}
    \end{subfigure}

    \centering
    \begin{subfigure}{0.49\linewidth}
        \centering
        \includegraphics[width=0.98\linewidth]{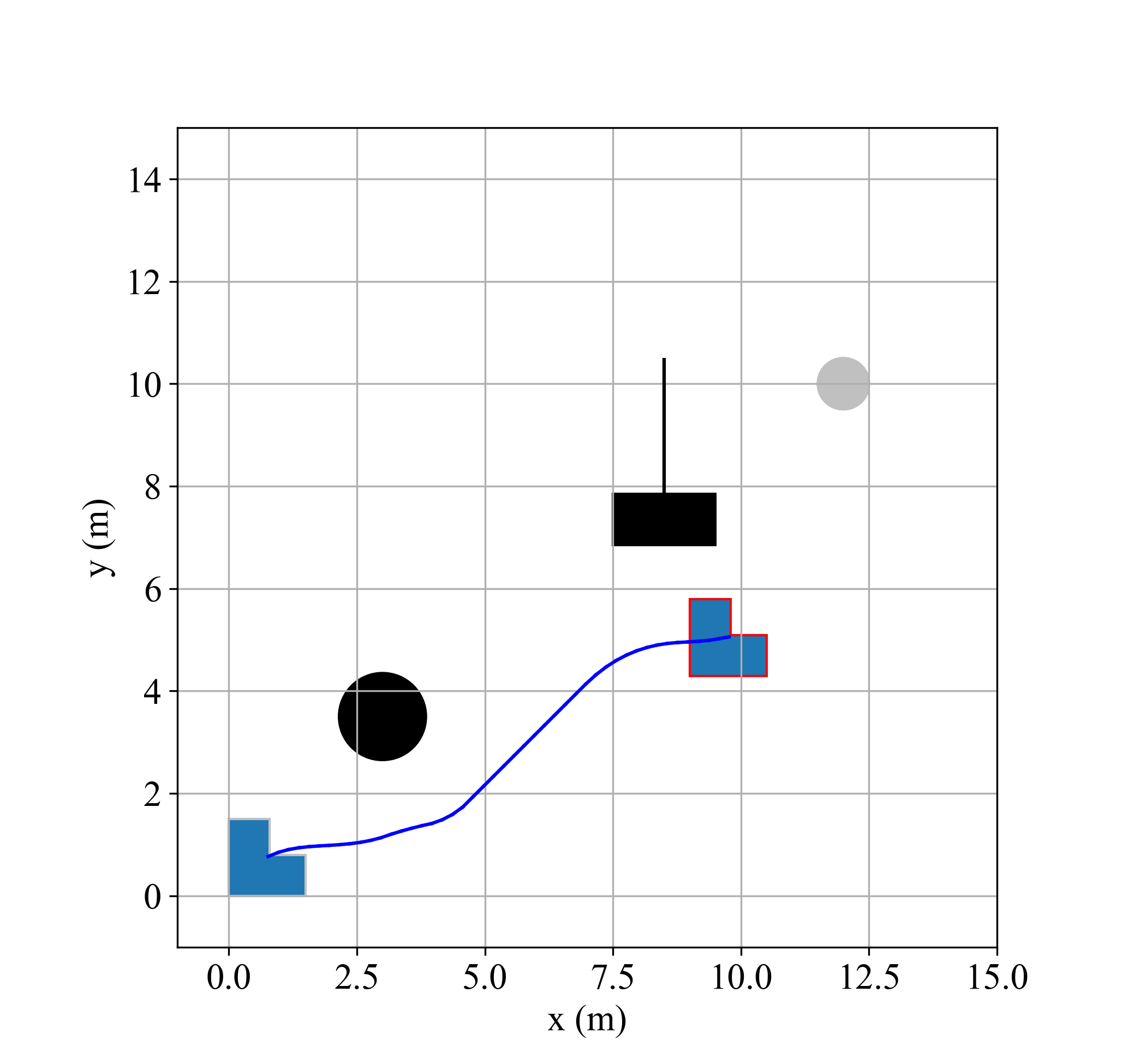}
        \caption{$t = 4.5 \, \si[per-mode=symbol]{\second}$}
        \label{subfig:integral_case3}
    \end{subfigure}
    \centering
    \begin{subfigure}{0.49\linewidth}
        \centering
        \includegraphics[width=0.98\linewidth]{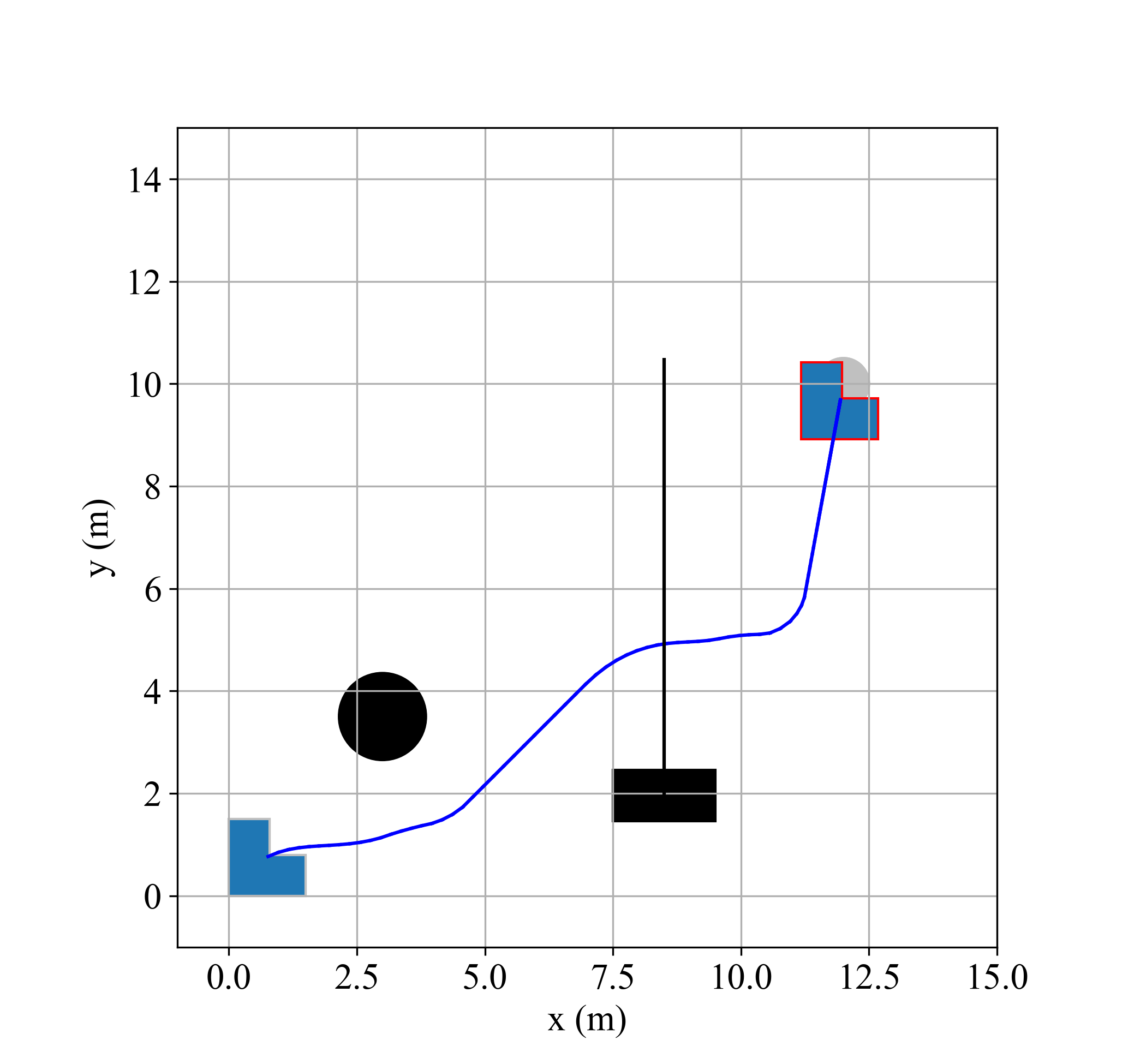}
        \caption{$t = 12.2 \, \si[per-mode=symbol]{\second}$}
        \label{subfig:integral_case4}
    \end{subfigure}
    \caption{Simulation results of navigating the robot with the single integral model to its destination, considering both static and dynamic obstacles. The destination of the L-shaped robot is denoted by a silver circle and its trajectory is denoted by the blue line. Both obstacles are represented in black, and the trajectory of dynamic obstacle is denoted by the black line.} 
    \label{fig:integral_robot}
\end{figure}
\begin{figure}
    \centering
    \begin{subfigure}{0.49\linewidth}
        \centering
        \includegraphics[width=0.95\linewidth]{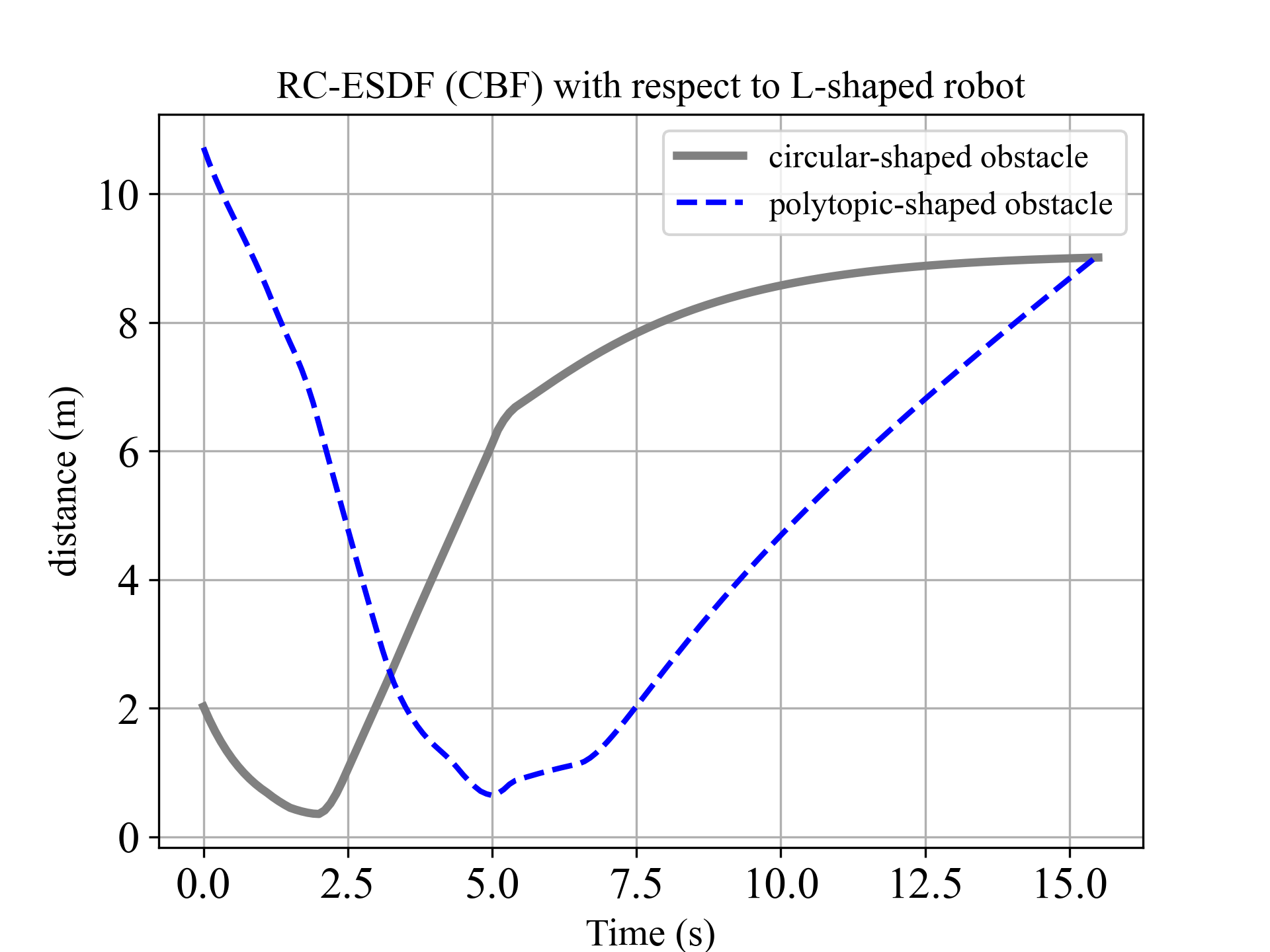}
        \caption{Changes in CBFs}
        \label{subfig:integral_cbf}
    \end{subfigure}
    \centering
    \begin{subfigure}{0.49\linewidth}
        \centering
        \includegraphics[width=0.95\linewidth]{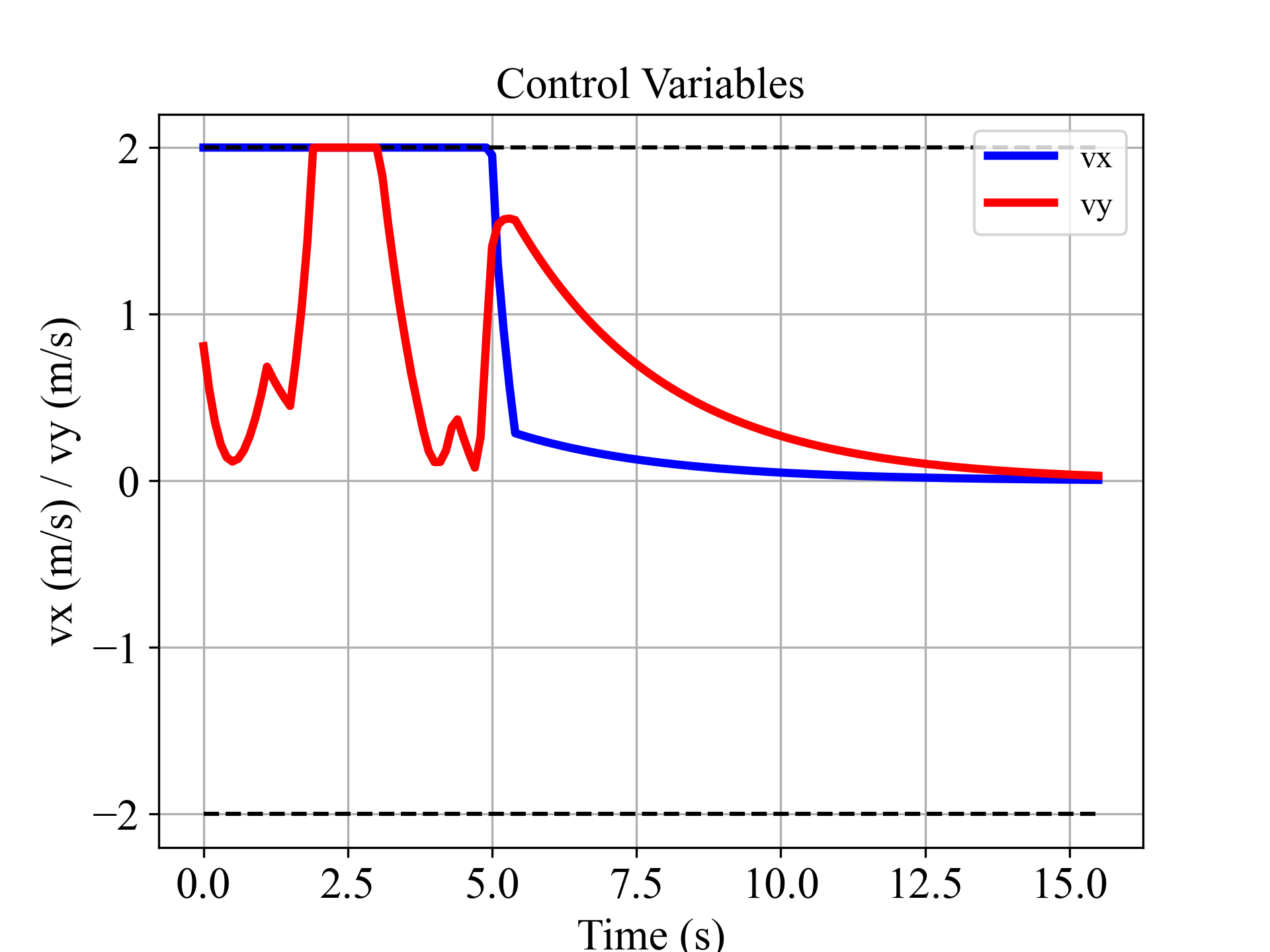}
        \caption{Changes in control variables}
        \label{subfig:integral_control}
    \end{subfigure}
    \caption{Changes over time in control barrier functions (CBFs) and control variables for the robot using the single integral model.} 
    \label{fig:integral_change}
\end{figure}
In this section, we demonstrate the effectiveness of our method for the robot using the single integral model.
The robot is in L-shaped, the initial and goal position of it are located in $(0.76 \, \si[per-mode=symbol]{\metre}, 0.76 \, \si[per-mode=symbol]{\metre})$ and $(12.0 \, \si[per-mode=symbol]{\metre}, 10.0 \, \si[per-mode=symbol]{\metre})$.
In addition, there are two obstacles in the environment: a static circular-shaped obstacle and a dynamic polytopic-shaped obstacle moved in a velocity of $(0.0 \, \si[per-mode=symbol]{\metre\per\second}, -0.7 \, \si[per-mode=symbol]{\metre\per\second})$.
Our approach can guide the L-shaped robot to its destination, while avoiding collisions with both static and dynamic obstacles, as shown in Fig.~\ref{fig:integral_robot}.
Since we use the RC-ESDF value to construct CBFs, any change in the RC-ESDF value represents a change in the CBFs.
In Fig.~\ref{subfig:integral_cbf}, we can observe that the robot avoids these two obstacles sequentially, and there is always a positive distance between the robot and the obstacles.
In Fig.~\ref{subfig:integral_control}, we show the changes of control variables, which always remain within the allowable range.
The boundary of the control variables are denoted by black dashed lines and here are same for both $v_x$ and $v_y$.
Our approach has an average computation time of $0.0116 \,\si[per-mode=symbol]{\second}$ making it suitable for real-time applications.

\subsection{Robot with Unicycle Model}
\begin{figure}
    \centering
    \begin{subfigure}{0.49\linewidth}
        \centering
        \includegraphics[width=0.98\linewidth]{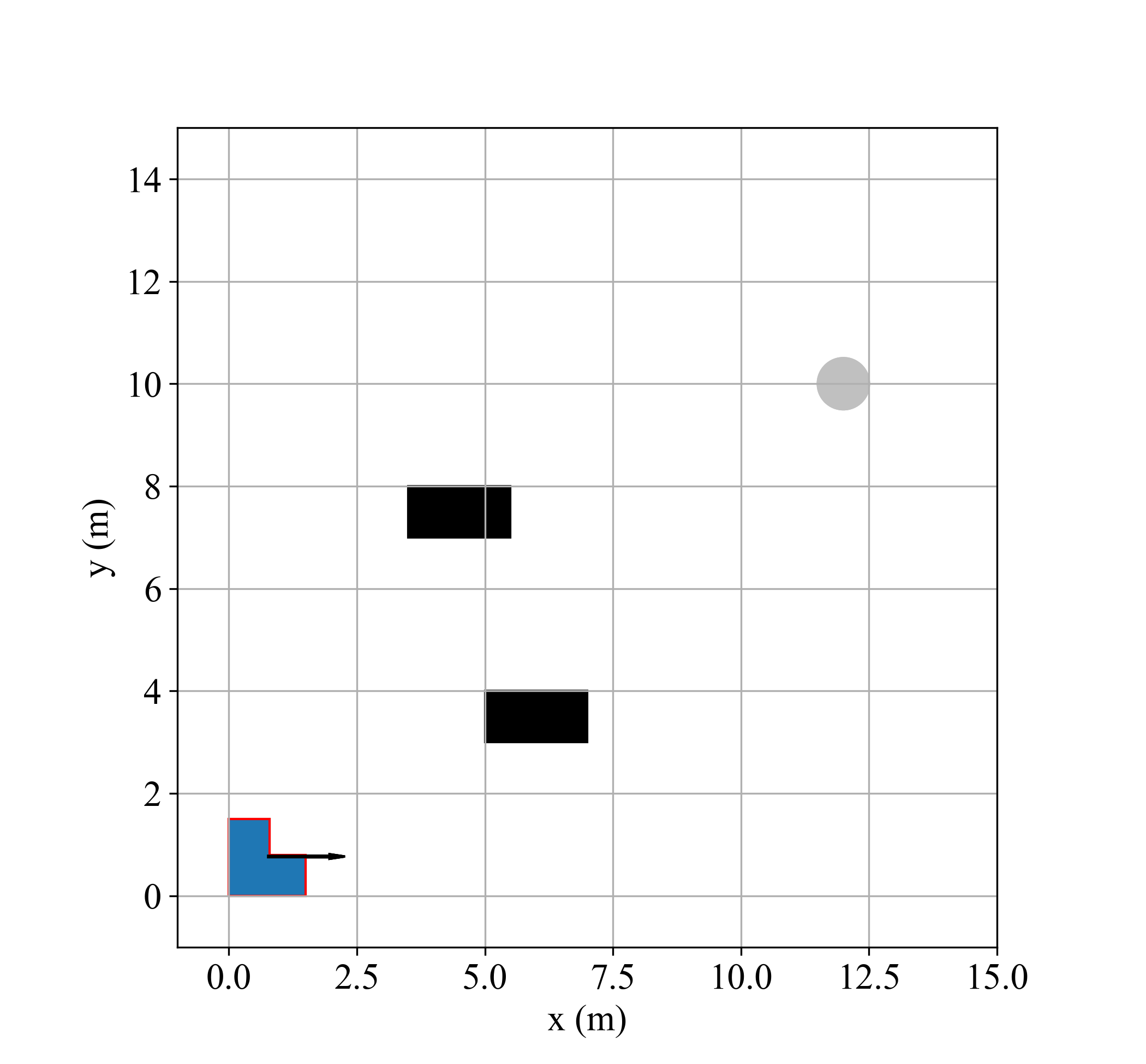}
        \caption{$t = 0.0 \, \si[per-mode=symbol]{\second}$}
        \label{subfig:unicycle_case1}
    \end{subfigure}
    \centering
    \begin{subfigure}{0.49\linewidth}
        \centering
        \includegraphics[width=0.98\linewidth]{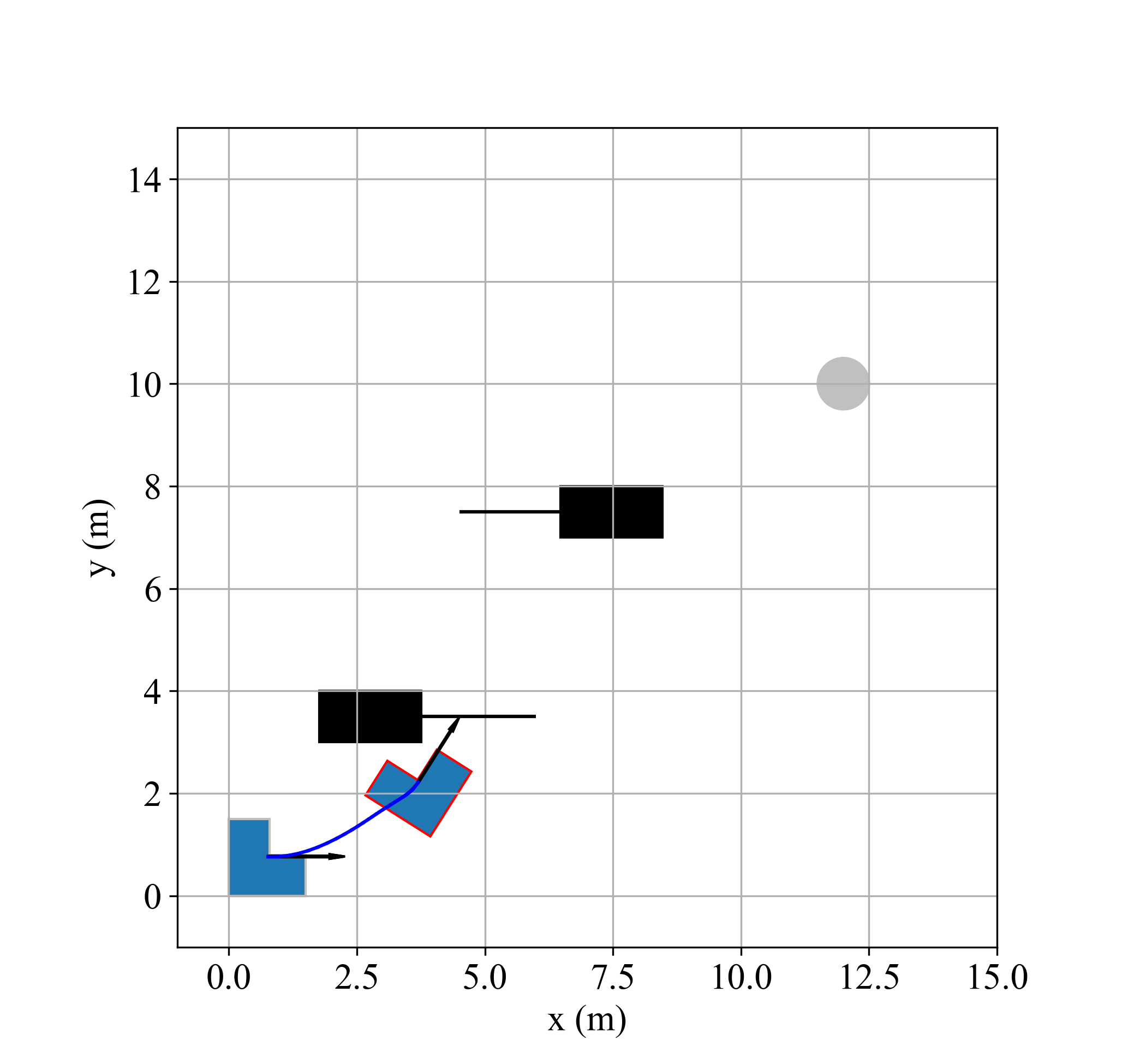}
        \caption{$t = 5.4 \, \si[per-mode=symbol]{\second}$}
        \label{subfig:unicycle_case2}
    \end{subfigure}

    \centering
    \begin{subfigure}{0.49\linewidth}
        \centering
        \includegraphics[width=0.98\linewidth]{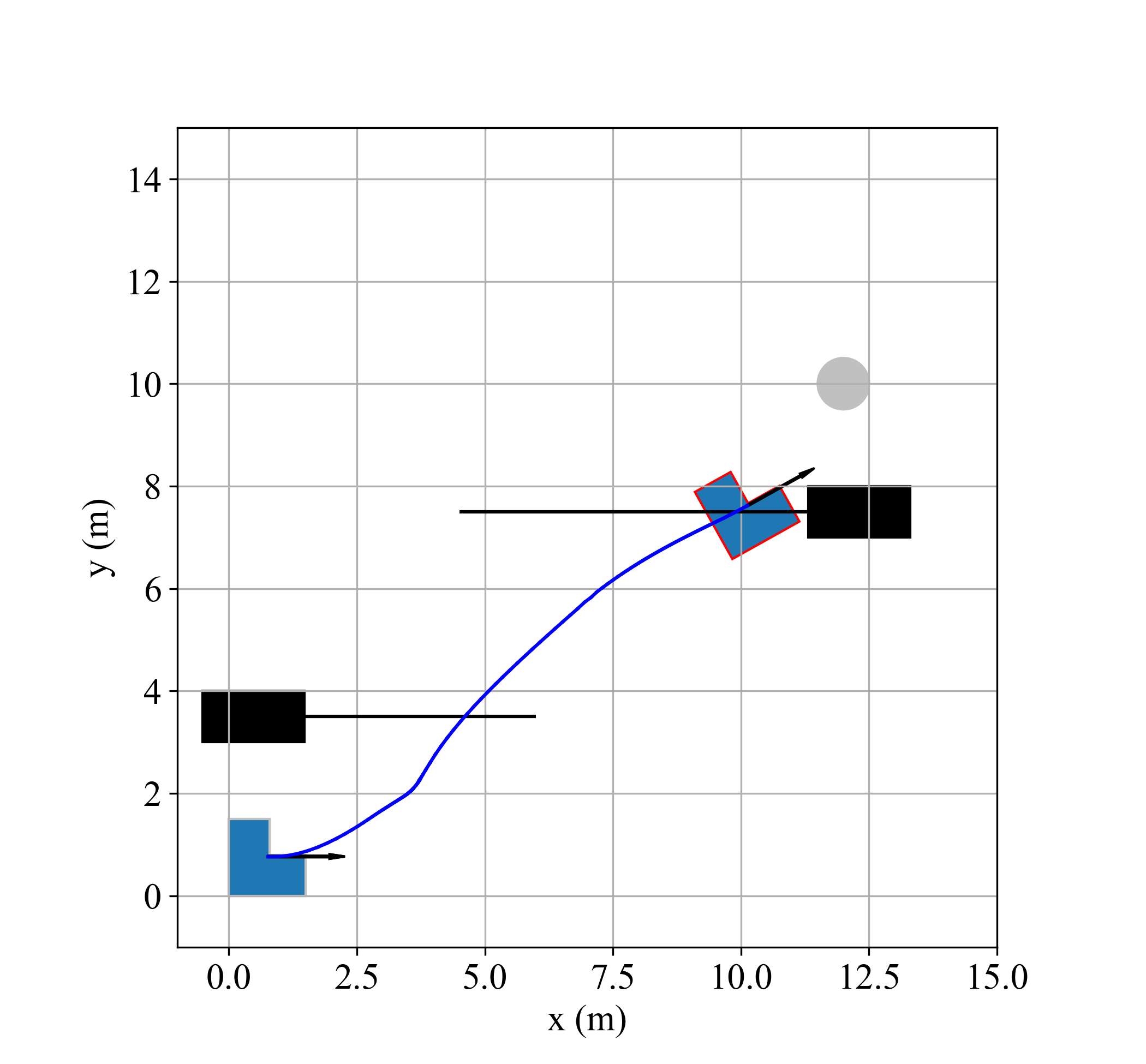}
        \caption{$t = 14.2 \, \si[per-mode=symbol]{\second}$}
        \label{subfig:unicycle_case3}
    \end{subfigure}
    \centering
    \begin{subfigure}{0.49\linewidth}
        \centering
        \includegraphics[width=0.98\linewidth]{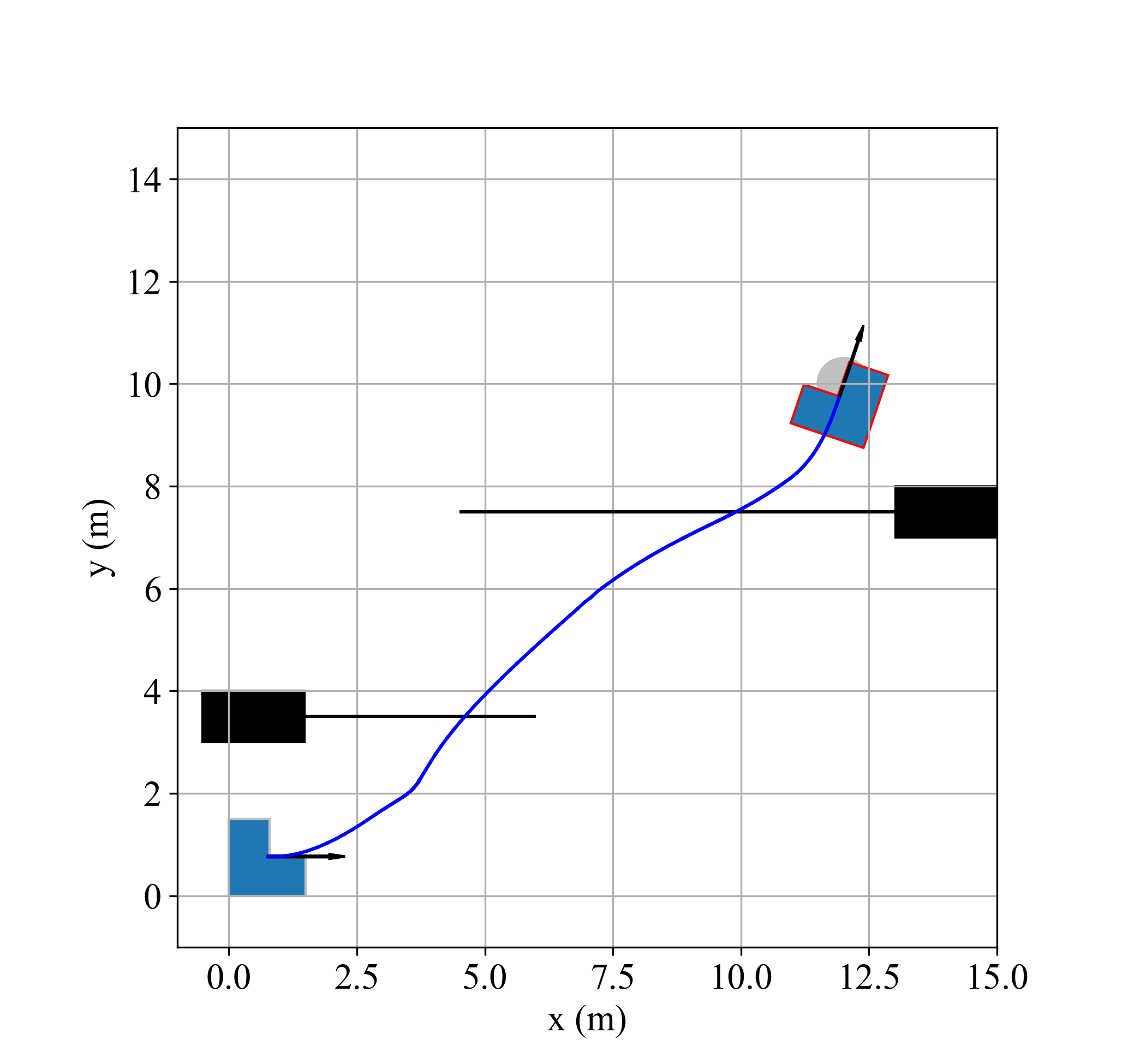}
        \caption{$t = 19.8 \, \si[per-mode=symbol]{\second}$}
        \label{subfig:unicycle_case4}
    \end{subfigure}
    \caption{Simulation results of navigating the robot with the unicycle model to its destination in the presence of dynamic obstacles. The destination of the L-shaped robot is denoted by a silver circle and its trajectory is shown as a blue line.
    Dynamic obstacles are depicted in black, with their trajectories indicated by black lines.}
    \label{fig:unicycle_robot}
\end{figure}

\begin{figure}
    \centering
    \begin{subfigure}{0.49\linewidth}
        \centering
        \includegraphics[width=0.95\linewidth]{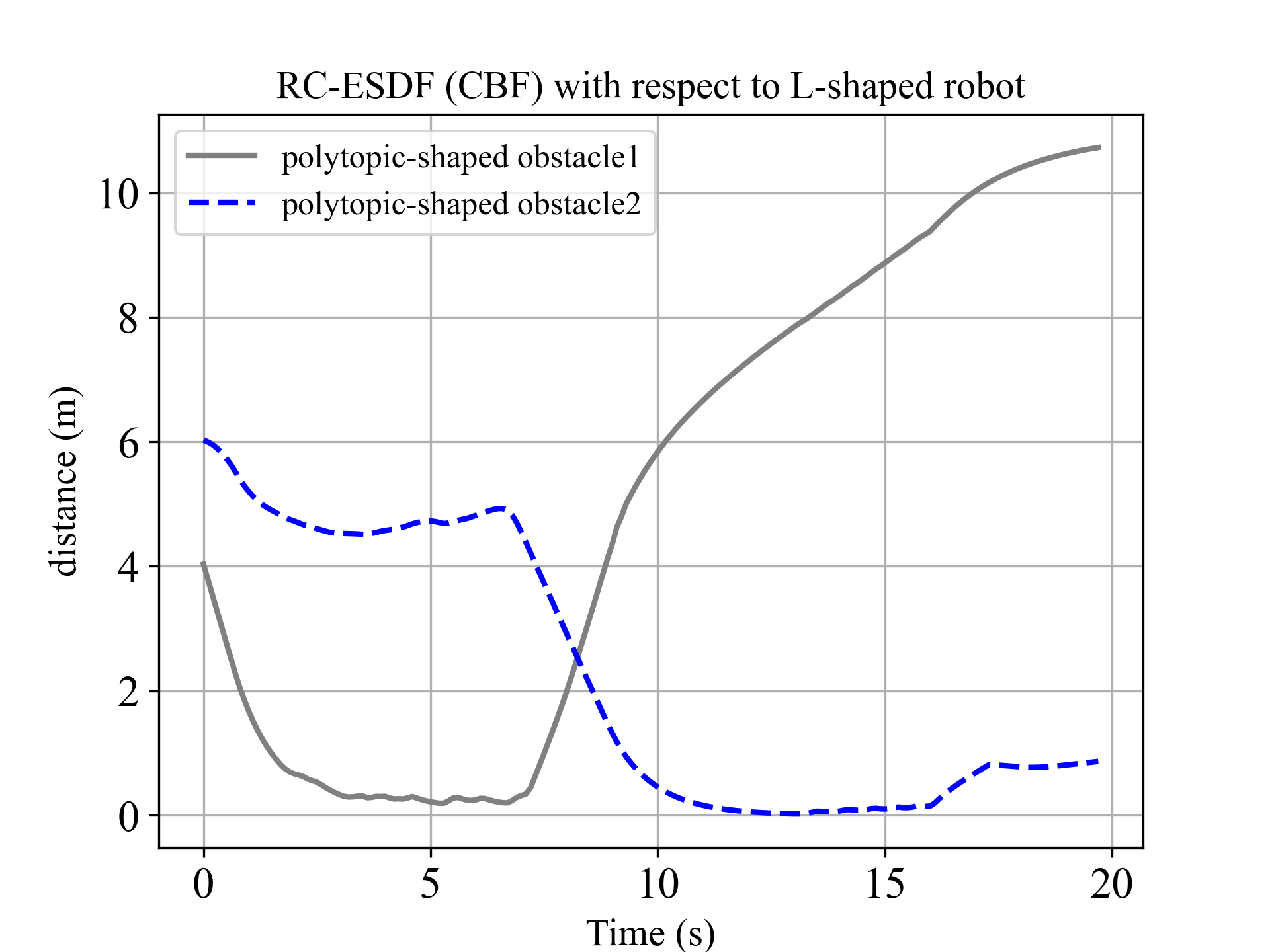}
        \caption{Changes in CBFs}
        \label{subfig:unicycle_cbf}
    \end{subfigure}
    \centering
    \begin{subfigure}{0.49\linewidth}
        \centering
        \includegraphics[width=0.95\linewidth]{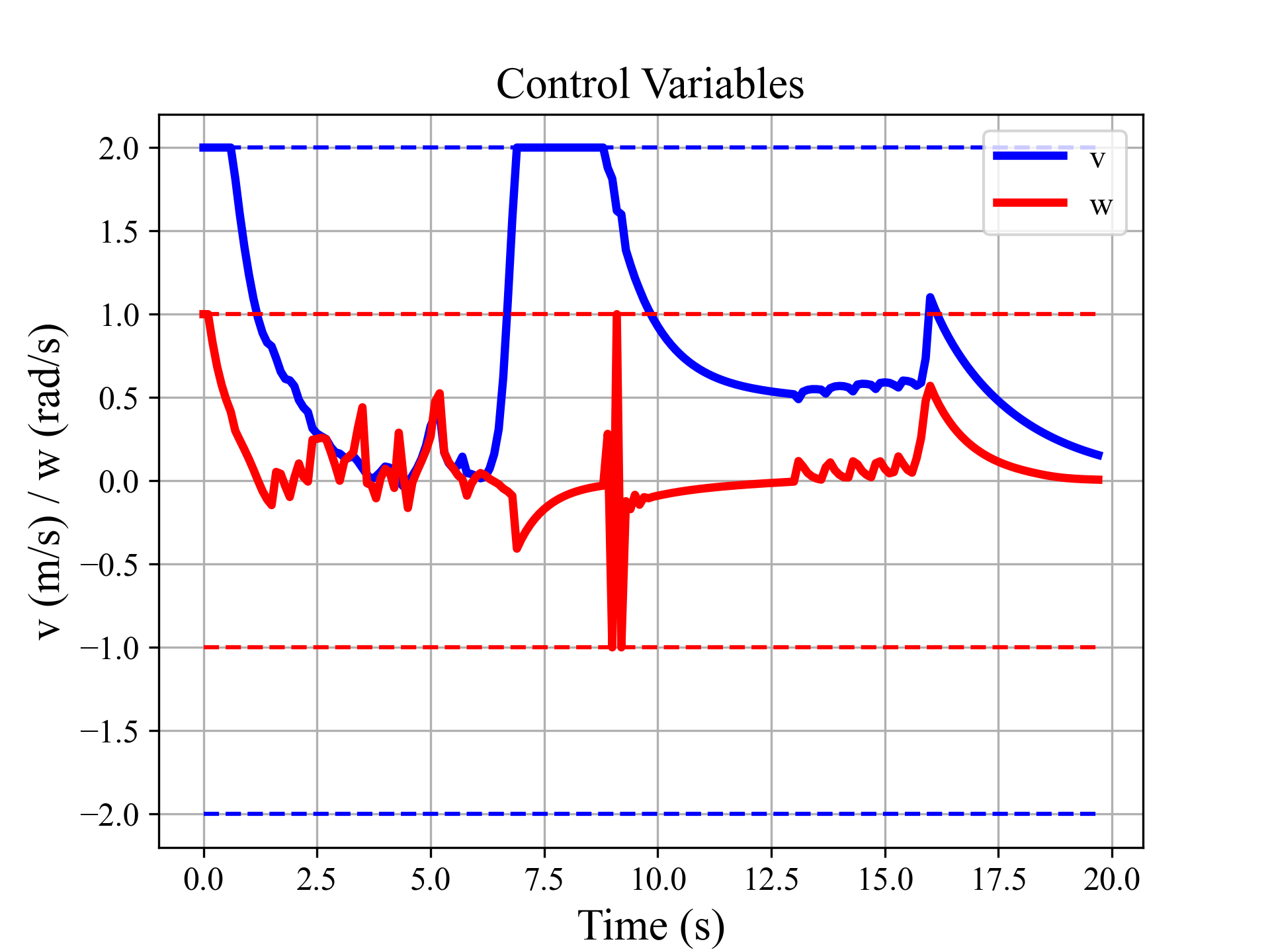}
        \caption{Changes in control variables}
        \label{subfig:unicycle_control}
    \end{subfigure}
    \caption{Changes over time in control barrier functions (CBFs) and control variables for the robot using the unicycle model.} 
    \label{fig:unicycle_change}
\end{figure}

In this section, we demonstrate the effectiveness of our approach for the robot using the unicycle model.
The robot is also in L-shaped, the initial and goal position are located in $(0.76 \, \si[per-mode=symbol]{\metre}, 0.76 \, \si[per-mode=symbol]{\metre})$ and $(12.0 \, \si[per-mode=symbol]{\metre}, 10.0 \, \si[per-mode=symbol]{\metre})$.
We set two dynamic polytopic-shaped obstacles in the environment, one travels from $(6.0 \, \si[per-mode=symbol]{\metre}, 3.5 \, \si[per-mode=symbol]{\metre})$ to $(0.5 \, \si[per-mode=symbol]{\metre}, 3.5 \, \si[per-mode=symbol]{\metre})$ at a speed of $(-0.6 \, \si[per-mode=symbol]{\metre\per\second}, 0.0 \, \si[per-mode=symbol]{\metre\per\second})$, the other travels from $(4.5 \, \si[per-mode=symbol]{\metre}, 7.5 \, \si[per-mode=symbol]{\metre})$ to $(14.0 \, \si[per-mode=symbol]{\metre}, 7.5 \, \si[per-mode=symbol]{\metre})$ at a speed of $(0.55 \, \si[per-mode=symbol]{\metre\per\second}, 0.0 \, \si[per-mode=symbol]{\metre\per\second})$.
The L-shaped robot with our proposed method can reach its destination while avoiding collision with dynamic obstacles, as shown in Fig.~\ref{fig:unicycle_robot}.
In Fig.~\ref{subfig:unicycle_cbf}, we observes that the robot successfully avoids the two polytopic-shaped robots, and maintains a distance greater than zero from the obstacles at all times.
The changes in the control variables $v$ and $\omega$ are shown in Fig.~\ref{subfig:unicycle_control}, which also satisfies the physical constraint of the robot, as indicated by the dashed line.
Our approach has an average computation time of $0.018 \,\si[per-mode=symbol]{\second}$ making it suitable for real-time applications.


\section{Conclusions}
\label{sec:conclusion}
In this paper, we propose a safety-critical controller that combines control Lyapunov functions (CLFs) and time-varying control barrier functions (time-varying CBFs) constructed using Robo-centric Euclidean Signed Distance Field (RC-ESDF) through a quadratic program (CLF-CBF-QP) to achieve navigation and whole-body collision avoidance with both static and dynamic obstacles.
We conducted several numerical simulations to validate the effectiveness of our proposed method under different robot dynamics. 
The results demonstrate that the proposed controller successfully navigates the robot to its destination while achieving whole-body collision avoidance with both static and dynamic obstacles.
However, there are limitations to our approach as its performance depends on the density of sampled collision points from the obstacle.
If the sampling points are sparse and the obstacle is relatively large, there may be instances where the robot passes through the interior of the obstacle, resulting in collisions.
Future works will focus on reducing the number of collision points considered in the optimization problem while maintaining an adequate sampling density.
Additionally, extending our proposed approach to multi-robot systems will be explored in future research.

{
\bibliographystyle{IEEEtran}
\bibliography{reference}

\begin{thebibliography}{10}
\providecommand{\url}[1]{#1}
\csname url@samestyle\endcsname
\providecommand{\newblock}{\relax}
\providecommand{\bibinfo}[2]{#2}
\providecommand{\BIBentrySTDinterwordspacing}{\spaceskip=0pt\relax}
\providecommand{\BIBentryALTinterwordstretchfactor}{4}
\providecommand{\BIBentryALTinterwordspacing}{\spaceskip=\fontdimen2\font plus
\BIBentryALTinterwordstretchfactor\fontdimen3\font minus \fontdimen4\font\relax}
\providecommand{\BIBforeignlanguage}[2]{{%
\expandafter\ifx\csname l@#1\endcsname\relax
\typeout{** WARNING: IEEEtran.bst: No hyphenation pattern has been}%
\typeout{** loaded for the language `#1'. Using the pattern for}%
\typeout{** the default language instead.}%
\else
\language=\csname l@#1\endcsname
\fi
#2}}
\providecommand{\BIBdecl}{\relax}
\BIBdecl

\bibitem{alonso2018cooperative}
J.~Alonso-Mora, P.~Beardsley, and R.~Siegwart, ``Cooperative collision avoidance for nonholonomic robots,'' \emph{IEEE Transactions on Robotics}, vol.~34, no.~2, pp. 404--420, 2018.

\bibitem{huang2023obstacle}
J.~Huang, Z.~Liu, J.~Zeng, X.~Chi, and H.~Su, ``Obstacle avoidance for unicycle-modelled mobile robots with time-varying control barrier functions,'' \emph{arXiv preprint arXiv:2307.08227}, 2023.

\bibitem{gilbert1988fast}
E.~G. Gilbert, D.~W. Johnson, and S.~S. Keerthi, ``A fast procedure for computing the distance between complex objects in three-dimensional space,'' \emph{IEEE Journal on Robotics and Automation}, vol.~4, no.~2, pp. 193--203, 1988.

\bibitem{zhang2020optimization}
X.~Zhang, A.~Liniger, and F.~Borrelli, ``Optimization-based collision avoidance,'' \emph{IEEE Transactions on Control Systems Technology}, vol.~29, no.~3, pp. 972--983, 2020.

\bibitem{han2021fast}
Z.~Han, Z.~Wang, N.~Pan, Y.~Lin, C.~Xu, and F.~Gao, ``Fast-racing: An open-source strong baseline for se(3) planning in autonomous drone racing,'' \emph{IEEE Robotics and Automation Letters}, vol.~6, no.~4, pp. 8631--8638, 2021.

\bibitem{li2021optimal}
B.~Li, Y.~Ouyang, Y.~Zhang, T.~Acarman, Q.~Kong, and Z.~Shao, ``Optimal cooperative maneuver planning for multiple nonholonomic robots in a tiny environment via adaptive-scaling constrained optimization,'' \emph{IEEE Robotics and Automation Letters}, vol.~6, no.~2, pp. 1511--1518, 2021.

\bibitem{wang2022geometrically}
Z.~Wang, X.~Zhou, C.~Xu, and F.~Gao, ``Geometrically constrained trajectory optimization for multicopters,'' \emph{IEEE Transactions on Robotics}, vol.~38, no.~5, pp. 3259--3278, 2022.

\bibitem{oleynikova2016signed}
H.~Oleynikova, A.~Millane, Z.~Taylor, E.~Galceran, J.~Nieto, and R.~Siegwart, ``Signed distance fields: A natural representation for both mapping and planning,'' in \emph{RSS 2016 workshop: geometry and beyond-representations, physics, and scene understanding for robotics}.\hskip 1em plus 0.5em minus 0.4em\relax University of Michigan, 2016.

\bibitem{geng2023robo}
S.~Geng, Q.~Wang, L.~Xie, C.~Xu, Y.~Cao, and F.~Gao, ``Robo-centric esdf: A fast and accurate whole-body collision evaluation tool for any-shape robotic planning,'' \emph{arXiv preprint arXiv:2306.16046}, 2023.

\bibitem{wang2023linear}
Q.~Wang, Z.~Wang, L.~Pei, C.~Xu, and F.~Gao, ``A linear and exact algorithm for whole-body collision evaluation via scale optimization,'' in \emph{2023 IEEE International Conference on Robotics and Automation (ICRA)}, 2023, pp. 3621--3627.

\bibitem{ames2014control}
A.~D. Ames, J.~W. Grizzle, and P.~Tabuada, ``Control barrier function based quadratic programs with application to adaptive cruise control,'' in \emph{53rd IEEE Conference on Decision and Control}, 2014, pp. 6271--6278.

\bibitem{ames2016control}
A.~D. Ames, X.~Xu, J.~W. Grizzle, and P.~Tabuada, ``Control barrier function based quadratic programs for safety critical systems,'' \emph{IEEE Transactions on Automatic Control}, vol.~62, no.~8, pp. 3861--3876, 2016.

\bibitem{wu2016safety}
G.~Wu and K.~Sreenath, ``Safety-critical control of a planar quadrotor,'' in \emph{2016 American control conference (ACC)}, 2016, pp. 2252--2258.

\bibitem{zeng2021safety}
J.~Zeng, B.~Zhang, and K.~Sreenath, ``Safety-critical model predictive control with discrete-time control barrier function,'' in \emph{2021 American Control Conference (ACC)}, 2021, pp. 3882--3889.

\bibitem{thirugnanam2022safety}
A.~Thirugnanam, J.~Zeng, and K.~Sreenath, ``Safety-critical control and planning for obstacle avoidance between polytopes with control barrier functions,'' in \emph{2022 International Conference on Robotics and Automation (ICRA)}.\hskip 1em plus 0.5em minus 0.4em\relax IEEE, 2022, pp. 286--292.

\bibitem{xiao2019control}
W.~Xiao and C.~Belta, ``Control barrier functions for systems with high relative degree,'' in \emph{2019 IEEE 58th conference on decision and control (CDC)}, 2019, pp. 474--479.

\bibitem{xiao2023barriernet}
W.~Xiao, T.-H. Wang, R.~Hasani, M.~Chahine, A.~Amini, X.~Li, and D.~Rus, ``Barriernet: Differentiable control barrier functions for learning of safe robot control,'' \emph{IEEE Transactions on Robotics}, vol.~39, no.~3, pp. 2289--2307, 2023.

\bibitem{jian2023dynamic}
Z.~Jian, Z.~Yan, X.~Lei, Z.~Lu, B.~Lan, X.~Wang, and B.~Liang, ``Dynamic control barrier function-based model predictive control to safety-critical obstacle-avoidance of mobile robot,'' in \emph{2023 IEEE International Conference on Robotics and Automation (ICRA)}, 2023, pp. 3679--3685.

\bibitem{ames2019control}
A.~D. Ames, S.~Coogan, M.~Egerstedt, G.~Notomista, K.~Sreenath, and P.~Tabuada, ``Control barrier functions: Theory and applications,'' in \emph{2019 18th European control conference (ECC)}, 2019, pp. 3420--3431.

\end{thebibliography}
}

\end{document}